\documentclass{ws-jcsc}

\usepackage{multirow}
\usepackage{booktabs}
\usepackage{tabularx}
\usepackage{makecell}
\usepackage{threeparttable}
\usepackage{arydshln}
\usepackage{mathrsfs}
\usepackage[title]{appendix}
\usepackage[section]{placeins}
\usepackage{etoolbox}
\AtBeginEnvironment{appendices}{%
  \counterwithin{figure}{section}%
  \counterwithin{table}{section}%
}
\usepackage{xcolor}
\usepackage[verbose]{hyperref}
\hypersetup{colorlinks=false,allbordercolors=blue,pdfborderstyle={/S/U/W 1}}

\makeatletter
\def\wsdraftnote{\today\quad\currenttime\qquad\qquad JCSC Submission}
\makeatother

\usepackage{overcite}

\renewcommand{\abstract}[1]{\begin{elsabstract}#1\end{elsabstract}}

\colorlet{revblue}{black}
\begin{document}

\markboth{X.~Wang et al.}%
{CogRAG: Stratified Retrieval and Reasoning for Heterogeneous Cognitive Demands}

\catchline{}{}{}{}{}
%%%%%%%%%%%%%%%%%%%%%%%%%%%%%%%%%%%%%%%%%%%%%%%%%%%%%%%%%%%%%%%%%%%%

\title{CogRAG: Tackling Heterogeneous Cognitive Demands in RAG via Stratified Retrieval and Reasoning}

\author{Xudong Wang$^{1}$}
\address{wangxvdong2019@gmail.com}

\author{Zilong Wang$^{3}$}
\address{zlw@idt.eitech.edu.cn}

\author{SU Kui$^{1,\dagger}$}
\address{suk@hzcu.edu.cn}

\author{Zhaoyan Ming$^{1,2,\dagger}$}
\address{mingcy@hzcu.edu.cn}

\address{$^{1}$School of Computer and Computing Science, Hangzhou City University,\\
Hangzhou, Zhejiang 310015, China\\
$^{2}$Innovation Center of Yangtze River Delta, Zhejiang University,\\
Jiaxing, Zhejiang 314100, China\\
$^{3}$Institute of Digital Twin, Eastern Institute of Technology,\\
Ningbo, Zhejiang 315200, China\\
$^{\dagger}$Joint corresponding authors.}

\maketitle

\abstract{Retrieval-Augmented Generation (RAG) frameworks typically process all queries through a one-size-fits-all pipeline, ignoring the heterogeneous cognitive demands of different tasks. This cognitive-blind approach causes two failure modes: cascading errors when low-level factual gaps trigger hallucinated reasoning, and reasoning-answer inconsistency in higher-order analytical tasks. We introduce CogRAG, a training-free, domain-agnostic framework that tackles these heterogeneous cognitive demands via stratified retrieval and reasoning. Inspired by Bloom's Taxonomy, CogRAG uses the predicted cognitive load of a query as a central control signal that coordinates two modules: Cognition-Adaptive Evidence Refinement supplements missing context via fact-centric or option-centric paths, and Cognition-Stratified Structured Reasoning replaces unconstrained chain-of-thought with cognition-aligned reasoning templates. We evaluate CogRAG on a demanding professional testbed, the Registered Dietitian qualification examination. CogRAG effectively reduces early-stage factual errors and eliminates reasoning-answer inconsistency, raising Qwen3-8B accuracy from 73.4\% to 85.8\% in single-choice mode and from 63.3\% to 80.5\% in scenario mode. These results highlight cognitive-stratified control as an effective, generalizable paradigm for reliable complex reasoning in large language models.}


%%%%% TODO1：不确定摘要中能否出现引号"one-size-fits-all"，还是说改成union

\section{Introduction}
\label{sec:introduction}

% 我的主线：
% 专业模型有效但训练成本高 → RAG 提供 training-free 知识注入 → 但 standard RAG 只解决“有没有知识”，没有解决“如何按认知层级使用知识” → Bloom’s Taxonomy 提供认知层级框架 → 当前模型缺少 cognitive control → 初步实验和错误分析证明 cognitive signal 有价值 → 引出后文 CogRAG。
%%%%%%% 我在这里说了我解决的是两个问题，一个是认知等级低级知识缺失，另一个是推理不一致问题。一个用RAG解决，一个用约束推理解决。后面的实验要有对应。

Retrieval-Augmented Generation (RAG) has emerged as a dominant paradigm for enhancing Large Language Models (LLMs) in complex question answering (QA) by grounding generated responses in external knowledge. However, while standard RAG significantly improves factual access, it typically treats all queries through a "one-size-fits-all" pipeline. Whether a user is asking for a basic definition or requesting a complex, multi-step diagnostic analysis, existing RAG systems apply the same flat retrieve-then-generate mechanism. This approach overlooks a fundamental characteristic of human expert problem-solving: \textbf{heterogeneous cognitive demands}. Different questions require entirely different cognitive operations, ranging from simple factual recall to complex rule application and critical evaluation. 

\begin{figure}[!htbp]
    \centering
    \includegraphics[width=\linewidth]{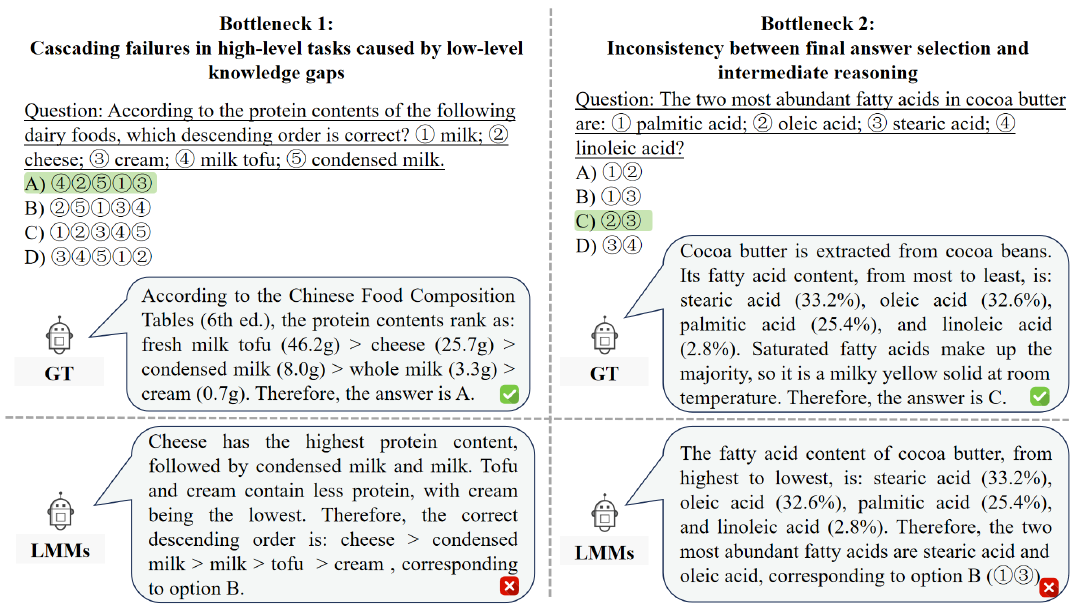}
    \caption{
    Representative bottlenecks in professional nutrition question answering. 
    Low-level factual errors can cascade into high-level reasoning failures, while reasoning-answer inconsistency may cause the model to derive correct evidence but select the wrong option.
    }
    \label{fig:intro_failure_modes}
\end{figure}

This "cognitive-blind" nature of current RAG frameworks leads to two major failure modes in complex reasoning tasks, as illustrated in Figure \ref{fig:intro_failure_modes}. First, \textbf{cascading errors from low-level factual gaps}: when a model fails to retrieve a foundational fact or concept (a lower-order cognitive demand), it often attempts to bridge the gap with hallucinated reasoning, causing the entire logical chain to collapse. Second, \textbf{reasoning-answer inconsistency in higher-order tasks}: when faced with questions requiring deep analysis or evaluation, standard unconstrained Chain-of-Thought (CoT) prompting often destabilizes. The model may generate a lengthy, seemingly plausible rationale, but ultimately select a final answer that contradicts its own intermediate reasoning. 

To overcome these bottlenecks, we draw inspiration from human cognitive science, specifically Bloom's Taxonomy \cite{krathwohl2002revision, anderson2001taxonomy}, which categorizes cognitive activities into a hierarchical structure (e.g., Remember, Understand, Apply, Analyze, Evaluate). Human experts do not merely accumulate facts; they dynamically adjust their information acquisition and logical deduction strategies based on the cognitive load of the problem. This suggests that reliable LLM reasoning requires not just retrieving evidence, but explicitly aligning the \textit{utilization} of that evidence with the cognitive operation demanded by the query.

Motivated by this, we propose \textbf{CogRAG}, a training-free, domain-agnostic framework designed to tackle heterogeneous cognitive demands via stratified retrieval and reasoning. Instead of a flat pipeline, CogRAG operates as a cognitive-driven control system. It first performs \textit{Cognitive Demand Profiling} to predict the cognitive load of the input query. Using this signal as a central controller, it dynamically coordinates two key phases:
\begin{romanlist}[(ii)]
    \item \textbf{Cognition-Adaptive Evidence Refinement}: Rather than relying on a static retriever, CogRAG evaluates initial evidence and triggers targeted refinement. For lower-order cognitive tasks, it executes a \textit{fact-centric} search to patch fundamental knowledge gaps; for higher-order tasks, it executes an \textit{option-centric} search to retrieve discriminative rules for complex comparisons.
    \item \textbf{Cognition-Stratified Structured Reasoning}: To eliminate logical disconnects, CogRAG replaces unconstrained CoT with structured reasoning templates tailored to specific cognitive levels. Lower-order tasks are constrained to direct evidence mapping, while higher-order tasks are forced into a rigorous "rule-application-comparison" deductive structure.
\end{romanlist}

{\color{revblue}This design directly follows from the two failure modes identified above. The predicted cognitive route serves as the bridge between motivation and implementation: it determines how missing or weak evidence should be refined, and how the resulting evidence should be used during reasoning. Specifically, \textbf{Cognition-Adaptive Evidence Refinement} targets \emph{cascading errors from low-level factual gaps} by repairing the evidence foundation before reasoning, while \textbf{Cognition-Stratified Structured Reasoning} targets \emph{reasoning--answer inconsistency} by enforcing cognition-aligned proof structures and final answer selection.}

To rigorously validate our framework, we require a testbed that features clear cognitive stratification, demands multi-step logical deduction, and has zero tolerance for hallucinations. Therefore, we evaluate CogRAG on a highly demanding professional testbed: the \textbf{Registered Dietitian qualification examination}. {\color{revblue}Experiments across different model families (Qwen3-8B and Llama-3.1-8B) show that CogRAG significantly reduces the onset of early-stage factual errors and, remarkably, eliminates reasoning-answer inconsistency, achieving 100\% logical alignment in valid responses.} Consequently, CogRAG raises the overall accuracy of Qwen3-8B from 73.4\% to 85.8\% in single-choice mode and from 63.3\% to 80.5\% in scenario-based mode.

The main contributions of this work are summarized as follows:
\begin{itemize}
    \item We identify a fundamental limitation in current RAG systems: the "cognitive-blind" approach to heterogeneous queries, which directly causes cascading factual errors and reasoning-answer inconsistencies in complex QA.
    \item We introduce CogRAG, a novel, domain-agnostic control framework that utilizes Bloom's Taxonomy as an inference-time signal to dynamically stratify both evidence acquisition and structured reasoning.
    \item We conduct comprehensive evaluations on a rigorous professional testbed. The results prove that cognitive-level guidance provides a highly effective, training-free paradigm for unlocking reliable complex reasoning in general-purpose LLMs.
\end{itemize}

\section{Related Work}
\label{sec:related_work}

\subsection{Advanced Retrieval-Augmented Generation}
% 修改说明：不要再写基础的BM25/Dense检索科普了（缩减），重点增加（Add）关于Adaptive RAG（自适应RAG）、Controlled/Structured RAG的文献，强调现有方法缺乏“认知层级”的控制。
Retrieval-Augmented Generation (RAG) improves language models by grounding generation in external evidence~\cite{lewis2020retrieval}. Beyond conventional sparse, dense, and hybrid retrieval methods~\cite{karpukhin2020dense,santhanam2022colbertv2,bruch2023analysis}, recent work has increasingly moved from fixed retrieve-then-generate pipelines toward adaptive retrieval mechanisms. For example, FLARE retrieves additional evidence during generation when future content is uncertain~\cite{jiang2023active}, Self-RAG introduces reflection signals to decide when and how retrieved evidence should be used~\cite{asai2024self}, Corrective RAG evaluates retrieval quality and triggers correction when the initial evidence is unreliable~\cite{yan2024corrective}, and Adaptive RAG routes queries according to task complexity~\cite{jeong2024adaptive}. These studies show that RAG benefits from dynamic control rather than a uniform retrieval strategy.

Another related direction is controlled or structured RAG, where external structures or constraints are introduced to regulate evidence organization and generation. Hierarchical and graph-based methods such as RAPTOR~\cite{sarthi2024raptor} and GraphRAG~\cite{edge2024local} organize retrieved knowledge into structured representations for multi-hop evidence aggregation, while constrained generation methods such as PICARD~\cite{scholak2021picard} and structured semantic parsing frameworks such as RESDSQL~\cite{li2023resdsql} improve output reliability by enforcing structural validity. However, existing adaptive and structured RAG methods mainly control retrieval timing, evidence granularity, or decoding format. They rarely model the cognitive demand of each query, nor do they use cognitive level as a global control signal to coordinate both retrieval and reasoning. CogRAG addresses this gap by introducing Cognitive Demand Profiling to guide evidence refinement and structured reasoning in a unified, training-free framework.

{\color{revblue}Table~\ref{tab:novelty_compare} compares CogRAG with representative adaptive or controlled RAG frameworks along several control dimensions. FLARE uses generation uncertainty to trigger active retrieval, CRAG uses retrieval-quality estimation to decide whether retrieved evidence should be corrected, Adaptive-RAG uses query complexity for routing, and Self-RAG uses reflection tokens to regulate retrieval and generation. CogRAG differs from these frameworks in its control signal and control scope: it uses predicted cognitive demand as an inference-time signal to coordinate both evidence refinement and reasoning-structure selection, followed by an explicit answer--reasoning consistency check.}

\begin{table}[!htbp]
\centering
\small
\setlength{\tabcolsep}{6pt}
\renewcommand{\arraystretch}{1.2}
\color{revblue}\caption{Comparison between CogRAG and representative RAG frameworks along control dimensions. Bold text highlights the proposed method and its cognitive-demand control signal.}
\label{tab:novelty_compare}
\begin{tabular}{@{}llccc@{}}
\toprule
\textbf{Method} & \textbf{Control signal}
 & \makecell{\textbf{Train-}\\\textbf{free}}
 & \makecell{\textbf{Ctrl.}\\\textbf{retrieval}}
 & \makecell{\textbf{Ctrl.}\\\textbf{reasoning}} \\
\midrule
FLARE~\cite{jiang2023active}          & generation uncertainty  & \checkmark & \checkmark & --         \\
CRAG~\cite{yan2024corrective}         & retrieval-quality score & \checkmark & \checkmark & --         \\
Adaptive-RAG~\cite{jeong2024adaptive} & query complexity        & \checkmark & \checkmark & --         \\
Self-RAG~\cite{asai2024self}          & reflection tokens       & --         & \checkmark & \checkmark \\
\textbf{CogRAG (ours)}                & \textbf{cognitive demand} & \checkmark & \checkmark & \checkmark \\
\bottomrule
\end{tabular}
\vspace{2pt}
\end{table}

\subsection{Cognitive Frameworks in LLM Reasoning}
\begin{figure}[t]
    \centering
    \includegraphics[width=0.95\linewidth]{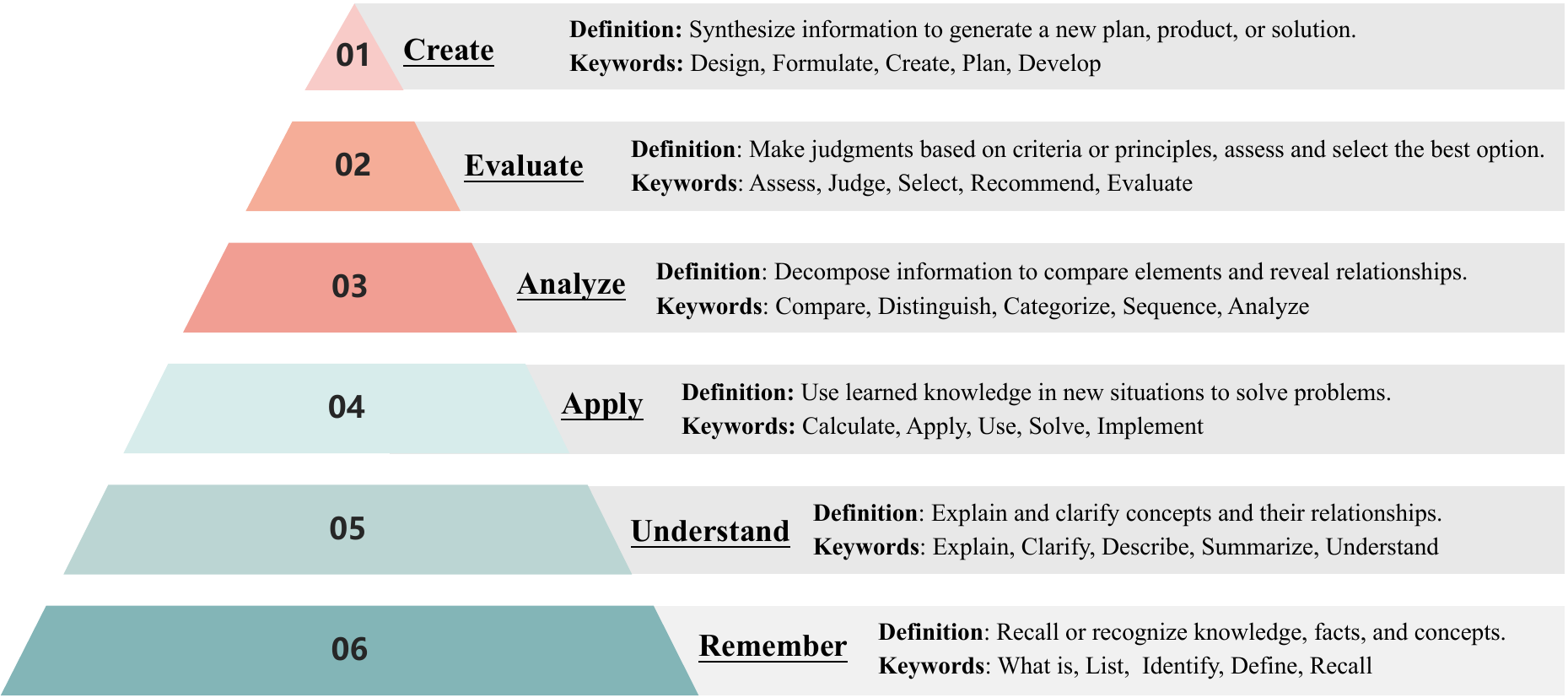}
    \caption{Illustration of Bloom's Taxonomy, which organizes cognitive activities from basic knowledge recall to higher-order reasoning and creative synthesis.}
    \label{fig:bloom-taxonomy}
\end{figure}

% 修改说明：将Bloom分类学融入到更广的“LLM认知推理”背景中。指出前人多用Bloom做“评测（Evaluation）”，而我们是首次将其作为RAG的“推理时控制信号（Inference-time control signal）”。
Bloom's Taxonomy is a classical framework in educational psychology that categorizes cognitive activities according to their complexity~\cite{anderson2001taxonomy}. The revised taxonomy defines six levels, namely Remember, Understand, Apply, Analyze, Evaluate, and Create, ranging from factual recall to higher-order reasoning and judgment. As illustrated in Figure~\ref{fig:bloom-taxonomy}, these levels form a progressive hierarchy. Since different levels imply different forms of knowledge use, Bloom's Taxonomy provides a principled basis for characterizing the heterogeneous cognitive demands faced by large language models.

Recent studies have introduced Bloom's Taxonomy into LLM evaluation, question answering, and educational applications. Existing work uses Bloom levels to diagnose benchmark coverage~\cite{huber2025llms}, construct cognitively annotated QA datasets and explanations~\cite{zoumpoulidi2025bloomxplain}, provide proximal context for zero-shot comprehension QA~\cite{sahu2021comprehension}, generate cognitively progressive follow-up questions~\cite{yadav2025recall}, or classify user queries into functional categories~\cite{samuel2025finder}. However, these studies mainly use cognitive levels for evaluation, annotation, question generation, or query analysis. In contrast, CogRAG treats cognitive demand as an inference time control signal. Through Cognitive Demand Profiling, each query is assigned a qualitative cognitive profile, which is then mapped into a compact route to globally guide retrieval refinement and structured reasoning.

\subsection{Complex Reasoning in Professional Domains}
% 修改说明 (大幅缩减/降级)：将原来的“领域特定QA”改写。不要列举太多FoodLMM之类的具体工作，而是论述“为什么专业领域（如医疗、法律）是验证复杂推理的最佳Testbed”。
Professional domains provide a rigorous testbed for evaluating complex reasoning in large language models, because they require factual recall, rule application, comparison among plausible alternatives, and judgment under specific constraints. These high-stakes and knowledge-intensive settings are particularly effective for revealing whether LLMs can perform reliable, evidence-grounded, and logically consistent reasoning beyond surface-level answer selection.
Existing studies show that professional QA often benefits from domain-aligned data and task-specific adaptation, as demonstrated by Med-PaLM in medical reasoning~\cite{singhal2025toward}, FinBen in financial evaluation~\cite{xie2024finben}, and food-oriented models such as FoodLMM~\cite{yin2025foodlmm} and FoodSky~\cite{zhou2025foodsky}. Rather than developing another domain-specialized model, this work uses the Registered Dietitian examination as a demanding professional QA testbed to evaluate a domain-agnostic control framework. Its mixture of lower-order factual questions and higher-order scenario-based questions enables us to examine whether Cognitive Demand Profiling can coordinate evidence refinement and structured reasoning across heterogeneous cognitive demands.

\section{Methodology: The CogRAG Framework}
\label{sec:method}

\subsection{Overview of Cognitive-Driven Control}
%%% 修改说明：重绘图3（Figure 3），画成一个带有“中枢控制器（Controller）”的系统架构图。强调这是一个领域无关（Domain-agnostic）的框架。现在的图3看起来像是一个单向流水线。建议改成**“控制系统”风格**。把“Bloom's Taxonomy / Cognitive Prediction”放在最上方作为Controller，画出两条虚线（控制信号）分别指向下方的“Retrieval”和“Reasoning”模块，体现出“认知信号全局指导”的理论高度。弱化TagRAG在图中的视觉占比。

Rather than treating retrieval and reasoning as a fixed sequential pipeline, CogRAG is designed as a \textbf{cognition-driven control framework}. Specifically, it first characterizes the cognitive demand of each query and then uses the resulting signal to coordinate downstream evidence acquisition and answer derivation. This design departs from the one-size-fits-all RAG paradigm by aligning both retrieval and reasoning with the cognitive operation required by the query.

{\color{revblue}This architecture is derived from the two failure modes of cognitive-blind RAG. First, since low-level factual gaps can cascade into later reasoning failures, the evidence foundation should be verified and repaired according to the cognitive demand of the query: lower-order queries require fact recovery, whereas higher-order queries require discriminative evidence for option comparison. This motivates \textbf{Cognition-Adaptive Evidence Refinement}. Second, since unconstrained chain-of-thought can lead to reasoning--answer inconsistency, the use of retrieved evidence should be constrained by cognition-aligned reasoning structures and checked by an explicit consistency safeguard. This motivates \textbf{Cognition-Stratified Structured Reasoning}.}

\begin{figure}[!htbp]
  \centering
  \includegraphics[width=1\linewidth,clip]{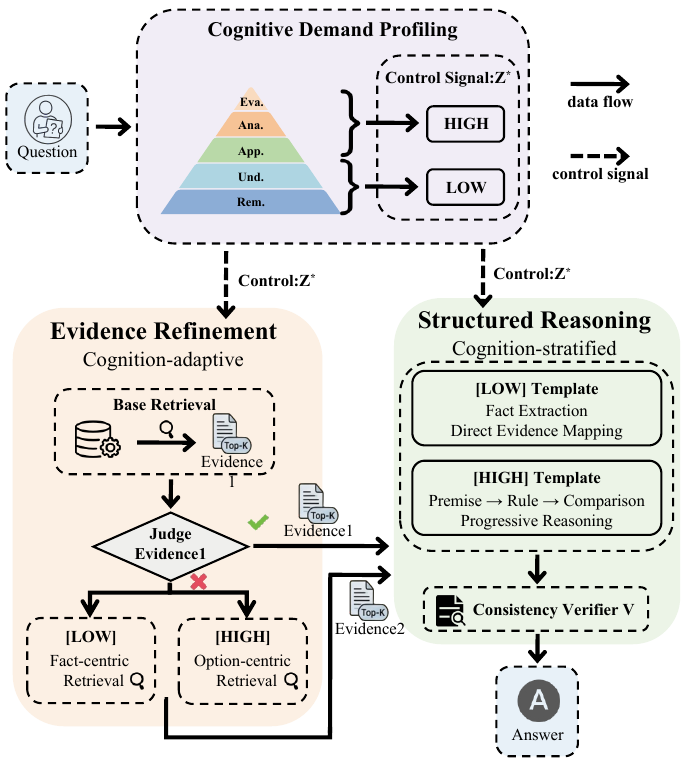}
  \caption{Architecture of CogRAG, a cognition-driven control framework. A central Cognitive Demand Profiler (top) estimates the cognitive load of the input query based on Bloom's Taxonomy and emits a binary route z*. This signal acts as a global control input (dashed arrows) that simultaneously governs two downstream modules: Cognition-Adaptive Evidence Refinement (left) and Cognition-Stratified Structured Reasoning (right). Solid arrows denote the data flow (query → evidence → answer), while dashed arrows denote the cognitive control flow.}
  \label{fig:Methods}
\end{figure}

As illustrated in Figure~\ref{fig:Methods}, the framework consists of one central controller and two cognition-controlled modules. The \textbf{Cognitive Demand Profiler} characterizes each input question through Bloom's Taxonomy and emits a compact binary route $z \in \{\text{LOW}, \text{HIGH}\}$, which serves as the global control signal. Conditioned on $z^{}$, \textbf{Cognition-Adaptive Evidence Refinement} dynamically adjusts the retrieval strategy to strengthen the evidence foundation, while \textbf{Cognition-Stratified Structured Reasoning} replaces unconstrained chain-of-thought with cognition-aware proof templates that align the final answer with the reasoning trace.

\subsection{Cognitive Demand Profiling}
%%% 修改说明：内容基本不变，但术语从“Prediction（预测）”升级为“Profiling（画像）”，强调这是在给输入问题做认知负荷的定性。
To enable cognition-level control throughout the CogRAG framework, we ground our method in Bloom's Taxonomy, which provides a structured basis for characterizing the qualitative cognitive demand of an input query. This profiling process distinguishes lower-order factual recall and comprehension from higher-order reasoning, comparison, and judgment. Since the Create level does not appear in our dataset, we define the cognitive-level space as:
\begin{equation}
\mathcal{L} = \{\text{Rem}, \text{Und}, \text{App}, \text{Ana}, \text{Eva}\},
\end{equation}

corresponding to Remember, Understand, Apply, Analyze, and Evaluate, respectively. Given an input query $x$ formed by concatenating the question and its options, the cognitive profiler assigns a qualitative cognitive-demand profile by estimating the most likely Bloom-level category through MAP inference:
\begin{equation}
\ell^{*} = \arg\max_{\ell \in \mathcal{L}} p_{\theta}(\ell \mid x).
\end{equation}

Although fine-grained five-level profiling provides interpretable cognitive characterizations, directly using these levels for downstream control can be unstable in practice, especially among adjacent higher-order categories. Motivated by the observation that lower-order tasks mainly require factual retrieval, whereas higher-order tasks require rule application, comparison, and multi-step reasoning, we consolidate the five Bloom levels into two broad cognitive routes, namely $\mathcal{L}_{\text{bin}} = \{\text{LOW}, \text{HIGH}\}$. The final cognitive route $z^{*}$ is obtained as follows:
\begin{equation}
z^{*} =
\begin{cases} 
\text{LOW}, & \text{if } \ell^{*} \in \{\text{Rem}, \text{Und}\}, \\ 
\text{HIGH}, & \text{if } \ell^{*} \in \{\text{App}, \text{Ana}, \text{Eva}\}.
\end{cases}
\end{equation}

This mapping from five Bloom levels to two cognitive routes preserves an interpretable intermediate cognitive profile while producing a compact and robust control signal for subsequent retrieval and reasoning modules.

A comparison of direct binary prediction and five-class prediction followed by binary mapping is presented in Appendix~\ref{secA2}.

The profiled cognitive route $z^{*}$ serves as a lightweight controller throughout the framework. It is used to select a cognition-specific system prompt $P_{\text{sys}}^{z^{*}}$ and is also injected into the user prompt to modulate reasoning depth and evidence utilization. Accordingly, the LLM output distribution is conditioned on three key factors: the input query $x$, the retrieved evidence block $E$, and the profiled cognitive route $z^{*}$:
\begin{equation}
p_{\psi}(y \mid x, E, z^{*}) = 
\text{LLM}_{\psi}\big(
\langle P_{\text{sys}}^{z^{*}},\, 
P_{\text{usr}}(x, E, z^{*}) \rangle
\big),
\end{equation}
where $P_{\text{sys}}^{z^{*}}$ denotes the route-specific system prompt, $P_{\text{usr}}(x, E, z^{*})$ denotes the user prompt composed from the query, retrieved evidence, and cognitive route, and $\langle\cdot,\cdot\rangle$ denotes the composition of the system and user prompts.

In the subsequent stages, Cognitive Demand Profiling provides the shared control signal $z^{*}$ for both evidence refinement and structured reasoning. Rather than merely assigning a category to the query, $z^{*}$ determines how evidence should be reinforced and how the reasoning process should be structured.

\subsection{Cognition-Adaptive Evidence Refinement}
A static base retriever often yields evidence that is either topically relevant but factually incomplete, or factually correct but insufficiently discriminative for complex choices. To mitigate the risk of reasoning over weak evidence, CogRAG introduces a dynamic feedback loop: \textbf{Cognition-Adaptive Evidence Refinement}(CAER). This module evaluates the initial retrieved evidence and, if deemed insufficient, triggers a targeted refinement process whose search strategy is strictly governed by the predicted cognitive route $z^*$.

This feedback mechanism is overseen by a lightweight Judge component $J$, which evaluates the alignment between the initial evidence $E_1$, the query $q$, and the options $O$. The Judge quantifies the "readiness" of the evidence using three core metrics:
\begin{itemize}
    \item \textbf{Relevance Score ($rel$).} A scalar value (0-100) assessing the overall thematic and factual coverage of the evidence regarding the query.
    \item \textbf{Support Values ($s$).} A vector $s = [s_A, s_B, s_C, s_D]$, where each element (0-100) represents the degree of evidence supporting a specific option. 
    \item \textbf{Target Label ($target$).} A categorical indicator specifying whether the question seeks the correct statement or an exclusionary exception (e.g., "which of the following is NOT...").
\end{itemize}

Based on the $target$ label, the Judge ranks the support scores to identify the top two leading candidate options, yielding their respective scores $s^{(1)}$ and $s^{(2)}$ (where $s^{(1)} \ge s^{(2)}$). The confidence margin $s^{(1)} - s^{(2)}$ serves as a critical indicator of how clearly the current evidence distinguishes the best answer from its closest competitor. The system triggers the refinement phase if the evidence fails either a Relevance Check (threshold $\alpha$) or a Confidence Margin Check (threshold $\beta$):
\begin{equation}
Trigger = \mathbb{I}(rel < \alpha \lor (s^{(1)} - s^{(2)}) < \beta)
\end{equation}

Once triggered, the refinement strategy is not arbitrary; it is explicitly routed by the cognitive demand $z^*$ to address the specific type of knowledge gap:

\vspace{1ex}
\noindent\textbf{Fact-Centric Path for LOW Cognitive Demands.} 
When $z^* = \text{LOW}$, the primary cognitive operations are \textit{Remember} and \textit{Understand}. Errors at this level typically stem from missing foundational entities, definitions, or basic concepts. Therefore, the refinement module executes a \textbf{fact-centric search}. It generates multiple expanded queries to aggregate broad general knowledge and definitions, ensuring a comprehensive coverage of the factual baseline. This prevents the model from hallucinating missing facts during the subsequent reasoning phase.

\vspace{1ex}
\noindent\textbf{Option-Centric Path for HIGH Cognitive Demands.} 
When $z^* = \text{HIGH}$, the cognitive operations shift to \textit{Apply}, \textit{Analyze}, and \textit{Evaluate}. For these tasks, the model usually possesses the basic facts but lacks the specific, discriminative rules required to adjudicate between highly plausible options. Consequently, the refinement module executes an \textbf{option-centric search}. By focusing exclusively on the top two competing candidates identified by the Judge, the module formulates targeted queries to retrieve comparative guidelines, specific boundary conditions, or exclusionary rules that can explicitly break the tie between the leading options.

Finally, the newly acquired context is fused with the initial evidence $E_1$ and filtered for semantic redundancy. This produces the optimized evidence block $E_2$, ensuring that the subsequent reasoning module receives highly discriminative, cognitively aligned context without exceeding the token budget.

\subsection{Cognition-Stratified Structured Reasoning}
While the previous module ensures the acquisition of high-quality evidence, how the LLM \textit{utilizes} this evidence is equally critical. In standard RAG pipelines, reasoning is typically elicited via unconstrained Chain-of-Thought (CoT) prompting. However, unconstrained CoT is "cognitive-blind": it often encourages verbose deduction for simple factual queries (leading to hallucination or "overthinking") and lacks rigorous structural boundaries for complex analytical tasks (leading to logical drift). This ultimately results in \textbf{reasoning-answer inconsistency}, where the model generates a plausible rationale but selects a final option that contradicts its own intermediate logic.

To enforce logical alignment, CogRAG introduces \textbf{Cognition-Stratified Structured Reasoning}(CSSR). Instead of allowing free-form CoT generation, this module requires the LLM (acting as a solver $G$) to construct a structured proof $P$ and derive a preliminary answer $\hat{a}$. The schema of this proof is dynamically dictated by the predicted cognitive route $z^*$, ensuring that the reasoning structure strictly matches the cognitive demand:
\begin{equation}
P, \hat{a} = G(q, O, E, z^*)
\end{equation}

To handle heterogeneous cognitive demands, we design two stratified proof templates, as summarized in Table~\ref{tab:cr_templates}.

\begin{table}[!htbp]
\centering
\caption{Structured proof templates for Cognition-Stratified Structured Reasoning (CSSR).}
\label{tab:cr_templates}
\footnotesize
\setlength{\tabcolsep}{3pt}
\renewcommand{\arraystretch}{1.05}
\begin{tabularx}{\linewidth}{@{}l l X@{}}
\toprule
\textbf{Mode} & \textbf{Field} & \textbf{Constraint} \\
\midrule
\multirow{4}{*}{\textbf{LOW}} 
& \texttt{key\_fact} & One-sentence core definition or classification. \\
& \texttt{evidence} & Concise points directly extracted from retrieved evidence. \\
& \texttt{elimination} & Brief rationale for excluding distractors, within three sentences. \\
& \texttt{answer} & Final option letter. \\
\midrule
\multirow{5}{*}{\textbf{HIGH}} 
& \texttt{assumptions} & Known conditions provided in the question. \\
& \texttt{rules} & 2--5 applicable professional guidelines or mechanisms. \\
& \texttt{application} & Application of rules to the given conditions. \\
& \texttt{comparison} & Evaluation of each option against the established rules. \\
& \texttt{answer} & Final option letter. \\
\bottomrule
\end{tabularx}
\end{table}

\vspace{1ex}
\noindent\textbf{LOW Schema (Direct Evidence Mapping).}
When $z^* = \text{LOW}$, the primary goal is factual recall or comprehension. Therefore, the LOW schema constrains the model to extract key facts directly from the retrieved evidence and produce only a concise elimination rationale. This avoids unnecessary multi-step deduction and reduces the risk of fabricating unsupported intermediate reasoning.

\vspace{1ex}
\noindent\textbf{HIGH Schema (Formal Deductive Logic).}
When $z^* = \text{HIGH}$, the task requires rule application, comparison, and judgment. Therefore, the HIGH schema enforces a rule-centric deductive structure that separates premises, applicable rules, rule application, and option comparison. This compartmentalized structure prevents free-form CoT from mixing evidence and conclusions in a tangled narrative, thereby maintaining a verifiable reasoning trace.

\vspace{1ex}
\noindent\textbf{Consistency Verification and Re-selection.}
To completely eliminate reasoning-answer inconsistency, CogRAG employs a lightweight verifier component $V$ to check the logical alignment between the generated proof $P$ and the preliminary answer $\hat{a}$:
\begin{equation}
Consistent = V(P, \hat{a})
\end{equation}

The verifier examines whether the final selected option logically follows from the \texttt{comparison} or \texttt{elimination} fields of the proof. If an inconsistency is detected (e.g., the proof explicitly invalidates Option B, but $\hat{a}$ is B), CogRAG triggers a deterministic re-selection mechanism. During re-selection, the model is prohibited from generating new rationales; it must derive the final answer \textit{strictly} based on the established rules and option comparisons already finalized in proof $P$. This constraint effectively closes the loop, guaranteeing that the final output is faithfully grounded in the cognitively aligned reasoning trajectory.

\section{Experimental Setup and Results}
\label{sec:exp}

\subsection{Professional QA Testbed and Experimental Setup}

\textbf{Testbed Selection.}
To rigorously evaluate the ability of CogRAG to handle heterogeneous cognitive demands, we select the Registered Dietitian (RD) qualification examination as our professional QA testbed. Unlike general-domain QA benchmarks, RD exam questions require both accurate domain knowledge and cognitively aligned reasoning. Some questions mainly test basic nutritional facts, definitions, and physiological mechanisms, while others involve scenario-based dietary intervention, rule application, and comparison among plausible clinical or nutritional decisions. Therefore, this benchmark provides a challenging and realistic environment for evaluating whether a RAG framework can adapt its retrieval and reasoning behavior to different cognitive demands.

The benchmark contains 811 single-choice questions and 379 scenario-based questions. Each question is annotated according to Bloom's taxonomy~\cite{krathwohl2002revision}, covering five cognitive levels: \textit{Remember}, \textit{Understand}, \textit{Apply}, \textit{Analyze}, and \textit{Evaluate}. The \textit{Create} level is not included because no item in the current benchmark requires open-ended solution generation. Each item is assigned the highest cognitive level required for solving the question, with initial labels generated by GPT-4o and then independently reviewed by three registered dietitians. The two question formats exhibit distinct cognitive distributions: single-choice questions are more concentrated on lower-order cognitive demands, with \textit{Remember} and \textit{Understand} accounting for 64.1\% of the items, whereas scenario-based questions require more higher-order reasoning, with \textit{Analyze} and \textit{Evaluate} accounting for 57.0\%. This distribution makes the RD exam an ideal stress test for cognition-driven retrieval and reasoning control.

A preliminary analysis of cognitive-level guidance is provided in Appendix~\ref{secA1}.

\textbf{Base Retriever Setup.}
To support the stringent domain-specific knowledge requirements of the RD testbed, we construct a nutrition-domain QA knowledge base as the foundation of retrieval-augmented generation. The corpus is built from FoodEarth, MedQA, and Nutri7Base through data curation, knowledge synthesis, and semantic deduplication. Specifically, Qwen3-Max is used to filter and refine nutrition-related entries from FoodEarth and MedQA, and to synthesize additional QA pairs from structured corpora in Nutri7Base. 

\begin{table}[!htbp]
\centering
\small
\setlength{\tabcolsep}{3.5pt} % 稍微缩小间距以容纳更多列
\renewcommand{\arraystretch}{1.2}
\caption{\textbf{Summary of the RAG Knowledge Base.} The corpus comprises 62,478 nutrition-specific QA pairs. Tags (T1-T6) represent our 6-category taxonomy for topic-aware retrieval.}
\label{tab:rag_kb}
\begin{tabular}{l r l cccccc}
\toprule
\textbf{Source} & \textbf{\# QA} & \textbf{Method} & \textbf{T1} & \textbf{T2} & \textbf{T3} & \textbf{T4} & \textbf{T5} & \textbf{T6} \\
\midrule
FoodEarth  & 25,629 & Curated     & 24817 & 0 & 812 & 0 & 0 & 0 \\
MedQA      & 12,652 & Curated     & 0 & 12652 & 0 & 0 & 0 & 0 \\
Nutri7Base & 24,197 & Synthetic$^\dagger$   & 0 & 0 & 14013 & 15444 & 10676 & 1219 \\
\midrule
Total & 62,478 & -- & 24817 & 12652 & 14825 & 15444 & 10676 & 1219 \\
\bottomrule
\multicolumn{9}{l}{\footnotesize $^\dagger$Generated via Qwen3-Max based on raw nutrition corpora.}
\end{tabular}
\end{table}

As summarized in Table~\ref{tab:rag_kb}, the final corpus contains 62,478 nutrition-specific QA pairs, each represented as a \{Question, Answer, Source, Tags\} tuple.

On top of this corpus, we implement Tag-Constrained Dense Retrieval as the base retriever. Instead of applying dense retrieval over the entire corpus, the base retriever first uses nutrition-domain metadata tags to restrict the candidate pool and then performs dense vector matching within the selected subset. The tag taxonomy covers six professional categories: dietary education, healthcare, food and nutrition, individual and group nutrition management, public nutrition and nutrition education, and catering management. This design adapts the retriever to the RD testbed by reducing topic drift and improving the domain relevance of retrieved evidence. CogRAG further builds upon this tag-constrained evidence foundation by introducing cognition-adaptive evidence refinement and cognition-stratified structured reasoning.

The comparison of BM25, dense, and hybrid retrieval that motivates our choice of dense retrieval is reported in Appendix~\ref{secA3}.

\textbf{Implementation Details and Baselines.}
We evaluate CogRAG using two representative 8B-scale instruction-tuned large language models: Qwen3-8B and Llama-3.1-8B. Both models are evaluated under the same prompt templates and inference settings to ensure a fair comparison across model families. Unless otherwise specified, we use deterministic decoding with temperature set to 0 and a maximum generation length of 256 tokens. For the Judge component in CogRAG, the relevance threshold $\alpha$ and the option-margin threshold $\beta$ are set to 50 and 35, respectively. These thresholds determine whether cognition-adaptive evidence refinement is triggered before final reasoning. All experiments are conducted on a server equipped with 4 NVIDIA RTX A6000 GPUs.

We compare CogRAG with several baselines. The non-retrieval baseline directly answers each question using the base LLM. For retrieval baselines, we include BM25\cite{robertson2009probabilistic}, dense retrieval, and hybrid retrieval. BM25 represents sparse lexical matching, dense retrieval uses BGE-M3 embeddings with FAISS indexing, and hybrid retrieval combines sparse and dense rankings through Reciprocal Rank Fusion. In addition, we include our tag-constrained base retriever as a stronger retrieval baseline to isolate the contribution of domain-aware evidence retrieval. By comparing CogRAG with these baselines, we examine whether cognition-adaptive evidence refinement and cognition-stratified reasoning provide additional gains beyond retrieval itself.

\subsection{Main Results on the Professional Testbed}

\begin{table}[!htbp]
\centering
\small
\setlength{\tabcolsep}{3.2pt}
\renewcommand{\arraystretch}{1.05}
\caption{Performance comparison on the RD exam benchmark. Bold and underlined values indicate the best and second-best results, respectively, for each metric within each model and question-format group. Tied results receive the same formatting.}
\label{tab:main_results}
\begin{tabular}{lll cc ccccc}
\toprule
 & & & \multicolumn{2}{c}{General Acc. (\%)} & \multicolumn{5}{c}{Cognitive Level Acc. (\%)} \\
\cmidrule(lr){4-5} \cmidrule(lr){6-10}
Model & Mode & Method & Overall & Macro & Rem. & Und. & App. & Ana. & Eva. \\
\midrule

% ========================= Qwen =========================
\multirow{12}{*}{Qwen} 
& \multirow{6}{*}{Single} 
& Baseline & 73.4 & 72.0 & 72.7 & 73.5 & 65.3 & 77.0 & 71.8 \\
& & BM25     & 72.1 & 71.0 & 69.8 & 72.7 & 66.3 & 74.6 & 71.8 \\
& & Dense    & 78.2 & 77.8 & 80.6 & 77.1 & 74.5 & 79.8 & 76.9 \\
& & Hybrid   & 71.9 & 70.0 & 69.8 & 73.4 & 62.2 & 75.4 & 69.2 \\
& & Base Ret.    & \underline{79.5} & \underline{80.3} & \underline{81.4} & \underline{77.8} & \underline{76.5} & \underline{81.0} & \underline{84.6} \\
& & CogRAG   & \textbf{85.8} & \textbf{87.3} & \textbf{90.7} & \textbf{84.6} & \textbf{84.7} & \textbf{84.1} & \textbf{92.3} \\
\cmidrule(lr){2-10}

& \multirow{6}{*}{Scenario} 
& Baseline & 63.3 & 60.5 & 73.3 & 55.4 & 45.5 & 72.3 & 56.0 \\
& & BM25     & 66.2 & 63.1 & 80.0 & 60.7 & 53.3 & 73.3 & 48.0 \\
& & Dense    & 72.6 & 70.6 & 86.7 & \underline{66.1} & 62.3 & 78.0 & \underline{60.0} \\
& & Hybrid   & 67.3 & 65.7 & 80.0 & 60.7 & 54.6 & 73.3 & \underline{60.0} \\
& & Base Ret.    & \underline{74.1} & \underline{72.0} & \underline{90.0} & 64.3 & \underline{66.2} & \underline{79.6} & \underline{60.0} \\
& & CogRAG   & \textbf{80.5} & \textbf{78.8} & \textbf{96.7} & \textbf{71.2} & \textbf{74.5} & \textbf{82.7} & \textbf{68.8} \\
\midrule

% ========================= Llama =========================
\multirow{12}{*}{Llama} 
& \multirow{6}{*}{Single} 
& Baseline & 49.3 & 49.3 & 40.6 & 53.1 & 46.9 & 49.2 & 56.4 \\
& & BM25     & 47.7 & 48.4 & 48.8 & 46.8 & 51.0 & 46.8 & 48.7 \\
& & Dense    & 58.1 & 58.4 & 59.7 & \textbf{61.4} & \underline{59.2} & 52.8 & 59.0 \\
& & Hybrid   & 56.4 & 57.4 & 57.4 & 57.7 & 57.1 & 53.2 & 61.5 \\
& & Base Ret.    & \underline{59.2} & \underline{60.8} & \underline{60.5} & \textbf{61.4} & 54.1 & \textbf{56.0} & \textbf{71.8} \\
& & CogRAG   & \textbf{60.3} & \textbf{62.7} & \textbf{68.2} & \underline{60.1} & \textbf{61.2} & \underline{54.8} & \underline{69.2} \\
\cmidrule(lr){2-10}

& \multirow{6}{*}{Scenario} 
& Baseline & 45.1 & 41.1 & 53.3 & 46.4 & 35.1 & 50.8 & 20.0 \\
& & BM25     & 47.0 & 47.8 & 56.7 & 46.4 & 45.5 & 46.6 & 44.0 \\
& & Dense    & 54.1 & 52.4 & 63.3 & \textbf{53.6} & 48.1 & 57.1 & 40.0 \\
& & Hybrid   & 51.7 & 51.1 & 63.3 & 48.2 & 45.5 & 54.5 & 44.0 \\
& & Base Ret.    & \underline{56.7} & \underline{56.5} & \underline{70.0} & 51.8 & \underline{54.6} & \textbf{58.1} & \underline{48.0} \\
& & CogRAG   & \textbf{57.8} & \textbf{58.9} & \textbf{72.9} & \underline{52.7} & \textbf{61.8} & \underline{57.8} & \textbf{49.5} \\
\bottomrule
\end{tabular}

\vspace{2pt}
\footnotesize
\emph{Note:} Base Ret. denotes the tag-constrained base retriever. Macro is the unweighted average over Rem., Und., App., Ana., and Eva.
\end{table}

Table \ref{tab:main_results} presents the overall performance of CogRAG compared to the baseline methods across both model families. CogRAG consistently achieves the best overall results, demonstrating its domain-agnostic effectiveness in enhancing LLM reasoning. Specifically, for Qwen3-8B, CogRAG elevates the accuracy from 73.4\% to 85.8\% on standard single-choice questions, and notably, from 63.3\% to 80.5\% on complex scenario-based questions. Similar substantial gains are observed with Llama-3.1-8B.

It is worth noting that standard retrieval-based methods do not always bring stable improvements on this professional testbed. Although dense retrieval generally provides useful external evidence, BM25 or hybrid retrieval may fail to outperform the non-retrieval baseline in several settings. For example, BM25 and Hybrid retrieval perform lower than the baseline in the Qwen3-8B single-choice setting. This suggests that simply adding retrieved evidence is insufficient for RD exam questions, where the model must not only access relevant knowledge but also apply it according to the cognitive demand of the question.

In contrast, CogRAG demonstrates its most significant performance margins precisely on these HIGH cognitive tasks. By dynamically routing the reasoning process into a rigorous, rule-centric deductive structure (the HIGH Schema), CogRAG effectively prevents the models from being overwhelmed by multi-step logical requirements. This substantial improvement in higher-order problem-solving empirically validates the necessity and efficacy of cognition-stratified control in complex RAG systems.

\subsection{Ablation Study}

\begin{table}[!htbp]
\centering
\small
\setlength{\tabcolsep}{4.5pt}
\renewcommand{\arraystretch}{1.2}
\caption{Ablation study on Qwen3-8B in the single-choice setting. Base Retriever denotes tag-constrained dense retrieval; CAER and CSSR denote the two proposed CogRAG modules. Bold values indicate the highest accuracy in each column.}
\label{tab:ablation_cograg}
\begin{tabular}{@{}ccc cc ccccc@{}}
\toprule
\multicolumn{3}{c}{Components} 
& \multicolumn{2}{c}{General Acc.(\%)} 
& \multicolumn{5}{c}{Cognitive Level Acc.(\%)} \\
\cmidrule(lr){1-3} \cmidrule(lr){4-5} \cmidrule(lr){6-10}
Base Retriever & CAER & CSSR 
& Overall & Macro 
& Rem. & Und. & App. & Ana. & Eva. \\
\midrule
 & & 
& 73.4 & 72.0 & 72.7 & 73.5 & 65.3 & 77.0 & 71.8 \\

\checkmark & & 
& 79.5 & 80.3 & 81.4 & 77.8 & 76.5 & 81.0 & 84.6 \\

\checkmark & \checkmark & 
& 82.9 & 83.5 & 83.0 & 82.6 & 78.6 & 83.7 & 89.7 \\

\checkmark & & \checkmark 
& 83.7 & 85.7 & 88.4 & 82.6 & 80.6 & 83.3 & 87.2 \\

\checkmark & \checkmark & \checkmark 
& \textbf{85.8} & \textbf{87.3} 
& \textbf{90.7} & \textbf{84.6} & \textbf{84.7} & \textbf{84.1} & \textbf{92.3} \\
\bottomrule
\end{tabular}
\end{table}

Table~\ref{tab:ablation_cograg} presents the component-wise ablation results on Qwen3-8B in the single-choice setting. Starting from the non-retrieval baseline, we progressively add the tag-constrained base retriever, \textbf{Cognition-Adaptive Evidence Refinement} (CAER), and \textbf{Cognition-Stratified Structured Reasoning} (CSSR). The results show that each component contributes to the final performance, and the complete CogRAG framework achieves the best overall and macro accuracy.

First, introducing the base retriever improves the overall accuracy from 73.4\% to 79.5\% and the macro accuracy from 72.0\% to 80.3\%. This indicates that restricting retrieval to domain-relevant tag subsets provides more useful evidence than relying solely on the parametric knowledge of the base model. The improvement is observed across all five cognitive levels, with especially clear gains in \textit{Apply} and \textit{Evaluate}, suggesting that domain-aware retrieval benefits both factual knowledge access and reasoning-oriented question answering.

Second, adding CAER further improves the overall accuracy to 82.9\%. Compared with the base retriever alone, the combination of base retrieval and CAER yields a 3.4-point gain in overall accuracy and improves all cognitive levels. The gains are particularly visible in \textit{Understand} and \textit{Evaluate}, where accuracy increases from 77.8\% to 82.6\% and from 84.6\% to 89.7\%, respectively. These results suggest that CAER strengthens the retrieved evidence when the initial retrieval results are insufficient or weakly discriminative.

Third, adding CSSR on top of the base retriever leads to an overall accuracy of 83.7\% and a macro accuracy of 85.7\%, outperforming the base retriever with CAER in macro-level performance. This shows that cognition-stratified reasoning constraints are especially useful for improving the consistency between the reasoning process and the final option selection. The improvement is substantial in \textit{Remember}, \textit{Understand}, and \textit{Apply}, indicating that CSSR not only benefits higher-order reasoning but also helps the model make more reliable use of retrieved factual evidence.

Finally, the full CogRAG framework, which combines the base retriever, CAER, and CSSR, achieves the best performance, with 85.8\% overall accuracy and 87.3\% macro accuracy. Compared with the base retriever alone, CogRAG improves overall accuracy by 6.3 points and macro accuracy by 7.0 points. It also achieves the best accuracy across all five cognitive levels. These results demonstrate that CAER and CSSR are complementary: CAER improves the quality and discriminative value of retrieved evidence, while CSSR improves the structured use of that evidence during answer selection.

{\color{revblue}
\subsection{Routing Robustness and Computational Overhead}
Since CogRAG relies on a predicted cognitive route, we first examine whether routing errors affect final QA accuracy. Table~\ref{tab:routing_robustness} compares different routing sources while keeping the remaining CogRAG pipeline unchanged. Predicted routing performs comparably to gold-label routing in both question formats. It achieves 85.8\% accuracy on single-choice questions, compared with 84.8\% under gold-label routing, while both settings achieve 80.5\% on scenario-based questions. These results indicate that routing prediction errors have only a limited impact on final QA accuracy. In contrast, deliberately mismatched routing strategies, such as random routing or forcing all questions to the LOW route, noticeably reduce performance on scenario-based questions, showing that appropriate cognitive routing is particularly important when higher-order reasoning is required.
}

\begin{table}[!htbp]
\centering
\begingroup
\color{revblue}
\small
\setlength{\tabcolsep}{4pt}
\renewcommand{\arraystretch}{1.1}
\caption{Effect of the routing source on final QA accuracy (Qwen3-8B, full CogRAG pipeline). $\Delta$ denotes the overall accuracy difference relative to the predicted route.}
\label{tab:routing_robustness}
\begin{tabular}{l cc c cc c}
\toprule
& \multicolumn{3}{c}{\textbf{Single}} 
& \multicolumn{3}{c}{\textbf{Scenario}} \\
\cmidrule(lr){2-4} \cmidrule(lr){5-7}
\textbf{Routing source} & Overall & Macro & $\Delta$ & Overall & Macro & $\Delta$ \\
\midrule
Predicted (default)    & 85.8 & 87.3 & --     & 80.5 & 78.8 & --     \\
Gold / oracle          & 84.8 & 85.9 & $-1.0$ & 80.5 & 79.8 & $0.0$  \\
Direct-binary          & 85.3 & 86.9 & $-0.5$ & 80.5 & 79.8 & $0.0$  \\
\midrule
Random                 & 84.6 & 85.9 & $-1.2$ & 78.4 & 77.8 & $-2.1$ \\
All-LOW (fact only)    & 85.7 & 86.8 & $-0.1$ & 78.1 & 78.0 & $-2.4$ \\
All-HIGH (option only) & 84.7 & 86.2 & $-1.1$ & 80.2 & 79.3 & $-0.3$ \\
\bottomrule
\end{tabular}
\endgroup
\end{table}

{\color{revblue}
Routing variations have a limited effect on single-choice questions, where all alternative settings remain within $1.2$ percentage points of the predicted route. Scenario-based questions are more sensitive: random routing and All-LOW routing reduce overall accuracy by $2.1$ and $2.4$ points, respectively, whereas All-HIGH routing causes only a $0.3$-point decrease. These results indicate that appropriate HIGH-route processing is particularly important for scenario-based questions requiring higher-order reasoning. Meanwhile, the overall pipeline remains relatively robust to moderate routing errors.
}

{\color{revblue}
We further examine whether finer-grained cognitive routing improves downstream control. To isolate the effect of routing granularity from cognitive-profiling errors, Table~\ref{tab:granularity} compares binary LOW/HIGH routing with five-level Bloom routing using gold cognitive labels, while keeping the remaining CogRAG pipeline and inference settings unchanged. Five-level routing causes only a small decrease on single-choice questions ($84.8\%\rightarrow84.2\%$), but substantially reduces accuracy on scenario-based questions ($80.5\%\rightarrow76.3\%$). The largest decrease is observed for Apply questions in the scenario-based setting ($83.1\%\rightarrow68.8\%$), suggesting that assigning Apply questions a separate level-specific control strategy does not improve performance on this benchmark. These results support the broader LOW/HIGH routing scheme as a more stable control granularity across the two question formats.

}

\begin{table}[!htbp]
\centering
\begingroup
\color{revblue}
\small
\setlength{\tabcolsep}{5pt}
\renewcommand{\arraystretch}{1.1}
\caption{Effect of routing granularity using gold cognitive labels (Qwen3-8B, full CogRAG pipeline). Rem./Und./App./Ana./Eva.\ denote per-level accuracies.}
\label{tab:granularity}
\begin{tabular}{l l c c c c c c}
\toprule
\textbf{Format} & \textbf{Granularity} & \textbf{Overall} & \textbf{Rem.} & \textbf{Und.} & \textbf{App.} & \textbf{Ana.} & \textbf{Eva.} \\
\midrule
\multirow{2}{*}{Single}
& 2-level & 84.8 & 90.7 & 84.3 & 78.6 & 83.7 & 92.3 \\
& 5-level & 84.2 & 86.8 & 82.9 & 82.7 & 84.1 & 89.7 \\
\midrule
\multirow{2}{*}{Scenario}
& 2-level & 80.5 & 83.3 & 66.1 & 83.1 & 82.7 & 84.0 \\
& 5-level & 76.3 & 93.3 & 64.3 & 68.8 & 80.1 & 76.0 \\
\bottomrule
\end{tabular}
\endgroup
\end{table}

{\color{revblue}
Finally, we report the computational overhead in Table~\ref{tab:overhead}. CogRAG adds judgment, structured reasoning, and verification steps, so it uses more LLM calls than standard prompting or direct Dense-RAG. However, it remains faster than standard CoT in wall-clock latency: CogRAG takes $11.04$\,s median and $14.79$\,s mean per query, compared with $14.15$\,s and $25.14$\,s for CoT. This is because CSSR uses a compact structured reasoning format, avoiding the long and unstable traces often produced by free-form CoT.
}

\begin{table}[!htbp]
\centering
\begingroup
\color{revblue}
\small
\setlength{\tabcolsep}{6pt}
\renewcommand{\arraystretch}{1.1}
\caption{Computational overhead per query (Qwen3-8B). Median and mean wall-clock latency are measured over $200$ single-choice queries on a single RTX A6000 with batch size $1$ and greedy decoding. The average call count excludes the lightweight cognitive-profiling call, which is made once per query; thus, full CogRAG uses about $4.1$ total LLM calls per query when profiling is included. Bold text and values highlight the full CogRAG configuration.}
\label{tab:overhead}
\begin{tabular}{l c c c}
\toprule
\textbf{Method} & \textbf{Avg.\ calls} & \textbf{Med.\ (s)} & \textbf{Mean (s)} \\
\midrule
Baseline (no retrieval)    & 1.0 & 0.39  & 0.40  \\
Baseline (retrieval)       & 1.0 & 0.55  & 0.56  \\
CAER only                  & 2.00      & 1.63  & 1.77  \\
CSSR only                  & 2.00      & 7.17  & 10.05 \\
CoT               & 2.07      & 14.15 & 25.14 \\
\textbf{CogRAG}            & \textbf{3.13} & \textbf{11.04} & \textbf{14.79} \\
\bottomrule
\end{tabular}
\endgroup
\end{table}

\subsection{In-depth Analysis: Resolving Cognitive Bottlenecks}
To further understand where the performance gains of CogRAG come from, we conduct an in-depth analysis around the two failure modes identified in the introduction: cascading factual errors and reasoning-answer inconsistency. The first analysis examines whether \textbf{Cognition-Adaptive Evidence Refinement} (CAER) reduces early-stage factual or conceptual gaps that may propagate into higher-order reasoning. The second analysis evaluates whether \textbf{Cognition-Stratified Structured Reasoning} (CSSR) improves the logical alignment between the generated reasoning trajectory and the final selected answer.

\subsubsection{Cognition-Adaptive Evidence Refinement Reduces Low-Level Cascading Errors}
A major vulnerability in complex QA is the phenomenon of \textit{cascading errors}, where a failure to retrieve a foundational fact (a LOW cognitive demand) forces the LLM to hallucinate premises, thereby corrupting the entire downstream reasoning chain. To quantify this failure mode, we analyze the \textit{Error Onset} distribution, which identifies the earliest cognitive step at which the model's reasoning first deviates from the ground truth.

\begin{figure}[!htbp]
  \centering
  \includegraphics[width=\linewidth]{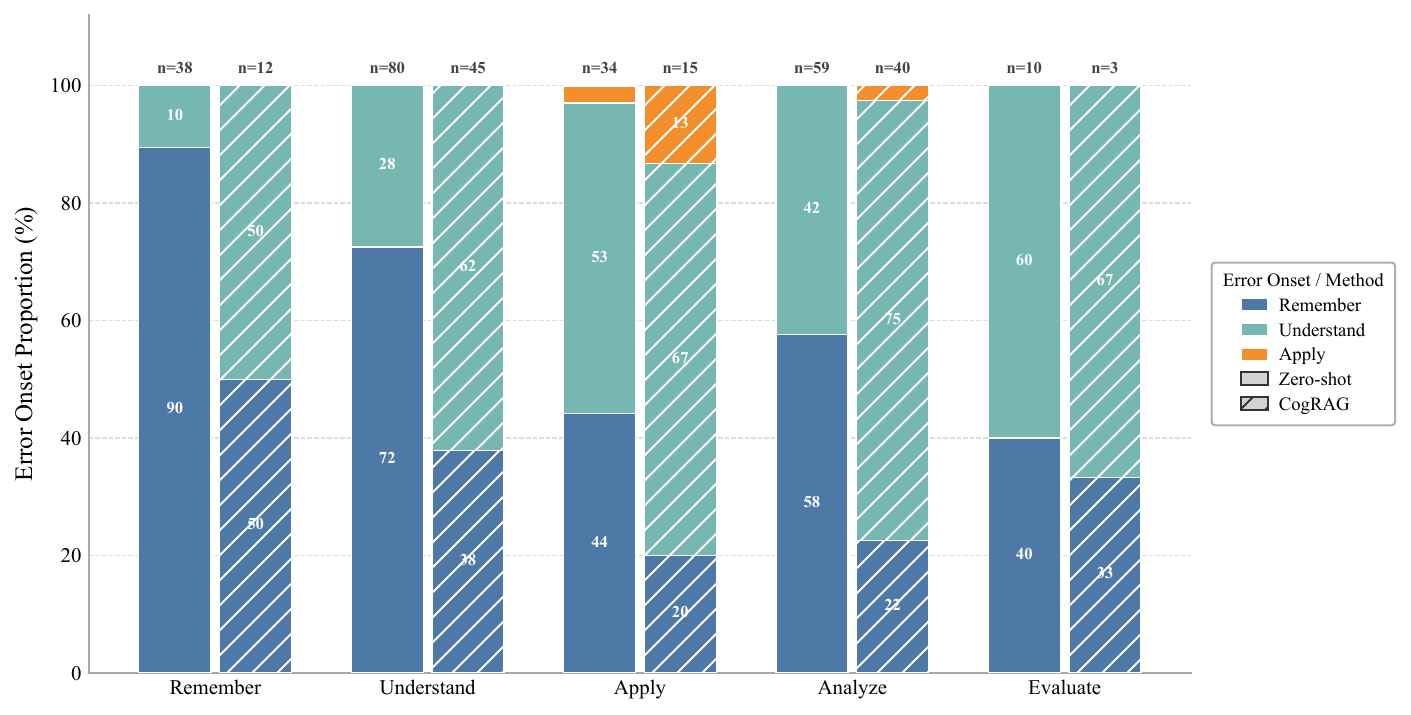}
  \caption{
  Distribution of error onset levels across target cognitive levels before and after CogRAG. Each group corresponds to the target cognitive level of the question, and stacked bars indicate the earliest cognitive stage at which the reasoning error begins.
  }
  \label{fig:error_onset}
\end{figure}

As shown in Figure~\ref{fig:error_onset}, under zero-shot inference, a large proportion of errors originate from lower-order cognitive stages, especially \textit{Remember} and \textit{Understand}. This indicates that even for higher-order questions, the model frequently fails before reaching the intended reasoning stage. In other words, many high-level reasoning failures are actually triggered by missing factual or conceptual evidence at earlier stages.

After applying CogRAG, the proportion of early-stage error onsets decreases substantially across most cognitive levels. For example, in \textit{Analyze} questions, the proportion of errors originating from \textit{Remember} decreases from 58\% to 22\%, while the total number of errors also decreases from 59 to 40. This corresponds to an approximate reduction from 34 to 9 \textit{Remember}-onset errors. Similar reductions can be observed in \textit{Remember}, \textit{Understand}, and \textit{Apply} questions. These results show that CogRAG effectively suppresses low-level factual failures before they cascade into later reasoning steps.

We further evaluate whether CAER strengthens the evidence foundation by comparing retrieval quality and downstream accuracy before and after adaptive refinement.

\begin{table}[!htbp]
\centering
\small
\caption{Effect of \textbf{Cognition-Adaptive Evidence Refinement} (CAER) on evidence quality and answer accuracy. Bold values indicate the higher value in each column.}
\label{tab:retrieval_performance}
\begin{tabular}{l c c c c}
\toprule
\multirow{2}{*}{\textbf{Method}} 
& \multicolumn{2}{c}{\textbf{LOW (Fact-Centric)}} 
& \multicolumn{2}{c}{\textbf{HIGH (Option-Centric)}} \\
\cmidrule(lr){2-3} \cmidrule(lr){4-5}
 & Rel. Score & Accuracy (\%) & Rel. Score & Accuracy (\%) \\
\midrule
Base Retriever      & 75.2 & 78.9 & 71.5 & 80.2 \\
 + CAER   & \textbf{77.8} & \textbf{82.7} & \textbf{74.8} & \textbf{83.0} \\
\bottomrule
\end{tabular}
\end{table}

As shown in Table~\ref{tab:retrieval_performance}, CAER improves both relevance scores and answer accuracy for LOW and HIGH tasks. For LOW, the fact-centric path increases the relevance score from 75.2 to 77.8 and accuracy from 78.9\% to 82.7\%, indicating better recovery of missing factual evidence. For HIGH, the option-centric path improves the relevance score from 71.5 to 74.8 and accuracy from 80.2\% to 83.0\%, suggesting that more discriminative option-level evidence supports higher-order reasoning. Together with Figure~\ref{fig:error_onset}, these results show that CAER reduces low-level knowledge gaps and prevents them from cascading into later reasoning stages.

\subsubsection{Cognition-Stratified Structured Reasoning Eliminates Reasoning-Answer Inconsistency}
The second major bottleneck in LLM reasoning is \textit{reasoning-answer inconsistency}. In higher-order cognitive tasks, standard unconstrained Chain-of-Thought (CoT) prompting may lead to a paradoxical outcome: the model generates a lengthy and seemingly plausible rationale, but ultimately selects a final option that is not fully supported by its own intermediate deductions. This problem occurs because free-form CoT lacks explicit structural boundaries, making the reasoning process vulnerable to logical drift during complex, multi-step comparisons.

To evaluate this failure mode, we measure the logical alignment between the generated rationale and the final selected answer. Specifically, we fix the retrieval stage as the base retriever and compare three reasoning strategies on top of the same retrieved evidence: direct answering, standard CoT, and the proposed \textbf{Cognition-Stratified Structured Reasoning} (CSSR).

\begin{table}[!htbp]
\centering
\small
\caption{Validity, reasoning-answer consistency, and accuracy comparison of different reasoning strategies under the same base retrieval setting. Direct denotes direct option selection based on the retrieved evidence. Unanswered denotes the proportion of outputs without a valid final option. Consistency measures the proportion of valid responses in which the final selected option is supported by the generated reasoning trajectory.}
\label{tab:consistency_metrics}
\begin{tabular}{l l c c c}
\toprule
Method & Level & Unanswered (\%) & Consistency (\%) & Accuracy (\%) \\
\midrule
\multirow{3}{*}{Base Retriever}
 & LOW    & 0.0 & -- & 78.9 \\
 & HIGH    & 0.0 & -- & 80.2 \\
 \cmidrule(lr){2-5}
 & Overall & 0.0 & -- & 79.5 \\
\midrule
\multirow{3}{*}{+ Standard CoT}
 & LOW    & 6.5 & 98.6 & 81.6 \\
 & HIGH    & 8.6 & 96.9 & 77.5 \\
 \cmidrule(lr){2-5}
 & Overall & 7.6 & 97.4 & 80.1 \\
\midrule
\multirow{3}{*}{+ CSSR}
 & LOW    & 2.4 & 100.0 & 84.4 \\
 & HIGH    & 0.5 & 100.0 & 83.1 \\
 \cmidrule(lr){2-5}
 & Overall & 1.4 & 100.0 & 83.7 \\
\bottomrule
\end{tabular}
\end{table}

As shown in Table~\ref{tab:consistency_metrics}, standard CoT does not consistently improve reasoning reliability. Although it slightly increases the overall accuracy from 79.5\% to 80.1\%, it also raises the unanswered rate to 7.6\%. This issue is more evident for HIGH tasks, where the unanswered rate reaches 8.6\% and the accuracy drops from 80.2\% to 77.5\%. These results indicate that unconstrained CoT may destabilize generation when complex evidence comparison and professional rule application are required.

CogRAG resolves this issue through CSSR. Instead of relying on free-form CoT, CSSR organizes the reasoning process into cognition-aligned proof schemas. For LOW cognitive demands, the model is guided to perform direct evidence mapping. For HIGH cognitive demands, the model is required to follow a more formal deductive structure, including explicit premise extraction, rule mapping, rule application, and structured option comparison. This design forces the model to compartmentalize its reasoning process and reduces the risk of logical drift.

Furthermore, the consistency verifier and deterministic re-selection mechanism provide an additional safeguard for answer selection. Compared with standard CoT, CSSR reduces the overall unanswered rate from 7.6\% to 1.4\% and improves the overall accuracy from 80.1\% to 83.7\%. More importantly, CSSR achieves a \textbf{100.0\% reasoning-answer consistency rate} among valid responses across both LOW and HIGH tasks. This result addresses the second bottleneck identified in the introduction, demonstrating that explicitly stratifying the reasoning structure according to cognitive demands can effectively align evidence, intermediate reasoning, and final answer selection.
\section{Conclusion}
\label{sec:conclusion}

{\color{revblue}
Although CogRAG shows strong performance, it is still evaluated within the specific setting of the Registered Dietitian qualification examination, which cannot fully represent more complex real-world scenarios such as open-ended consultation and multi-turn interaction. Within this setting, CogRAG consistently outperforms standard retrieval baselines across two model families, reduces cascading factual errors, and removes the observed reasoning--answer inconsistency, achieving $100.0\%$ consistency among valid responses. In future work, we will extend cognitive-level control beyond multiple-choice exam settings to open-ended nutrition consultation QA. In particular, we will explore cognitive-level-guided retrieval strategies, so that the system can better adapt evidence acquisition and answer generation to different user needs.

}

\section*{Acknowledgments}
This research was funded by the National Key Technologies Research and Development Program in China under Grant No. 2023YFF1104202-02. This research was also supported by the advanced computing resources provided by the Supercomputing Center of Hangzhou City University.

% \section*{Conflict of Interest}
% The authors declare that they have no known competing financial interests or personal relationships that could have appeared to influence the work reported in this paper.

% \section*{Data Availability}
% The datasets generated during and/or analysed during the current study are available from the corresponding author on reasonable request.

\section*{Code Availability}
{\color{revblue}The code used in this study is publicly available at: \url{https://github.com/SandyCrystal/CogRAG}.}

\begin{appendices}
\renewcommand{\thefigure}{\Alph{section}.\arabic{figure}}
\renewcommand{\thetable}{\Alph{section}.\arabic{table}}
\setcounter{figure}{0}
\setcounter{table}{0}

\section{Cognitive-Level Prediction for Routing}
\label{secA2}
%%% 认知路由命中率比较
The premise of cognitive-level injection is the reliable prediction of the cognitive level required by each question. To this end, we compare two routing strategies: direct binary prediction (Direct 2-way) and five-class prediction followed by binary mapping (5-way-to-2-way Mapping). As shown in Table~\ref{tab:routing_accuracy}, Direct 2-way prediction exhibits a clear category calibration bias under both zero-shot and few-shot settings: it achieves a high hit rate on low-order questions but performs poorly in identifying high-order tasks. In contrast, 5-way-to-2-way Mapping achieves better overall performance and more balanced recognition across categories, reaching an overall hit rate of 81.0\% under the few-shot setting. Therefore, we adopt the few-shot five-class mapping strategy as the default cognitive-level prediction method in the subsequent experiments.

\begin{table}[!htbp]
\centering
\small
\caption{Comparison of routing hit rates between direct binary prediction and five-class-to-binary mapping.}
\label{tab:routing_accuracy}
\begin{tabular}{l ccc ccc}
\toprule
\multirow{2}{*}{\textbf{Setting}} 
& \multicolumn{3}{c}{\textbf{Direct 2-way}} 
& \multicolumn{3}{c}{\textbf{5-to-2 Mapping}} \\
\cmidrule(lr){2-4} \cmidrule(lr){5-7}
& Overall & Low & High & Overall & Low & High \\
\midrule
Zero-shot & 58.1 & 98.1 & 14.7 & 67.8 & 63.0 & 73.0 \\
Few-shot  & 63.4 & 96.7 & 27.3 & 81.0 & 87.2 & 74.3 \\
\bottomrule
\end{tabular}
\end{table}

\section{Preliminary Analysis of Cognitive-Level Guidance}
\label{secA1}
%%% 认知等级对于解题是有帮助的。
%%% 基础模型在各认知等级上的准确率对比
%%% 认知等级注入方法的准确率对比

To examine whether cognitive-level information is useful for professional nutrition question answering, we first conduct a preliminary analysis on the baseline model. Specifically, we analyze its performance across question formats and cognitive levels, and then evaluate whether explicitly injecting cognitive-level information into the prompt can improve answer accuracy.

\begin{figure}[!htbp]
  \centering
  \includegraphics[width=\linewidth]{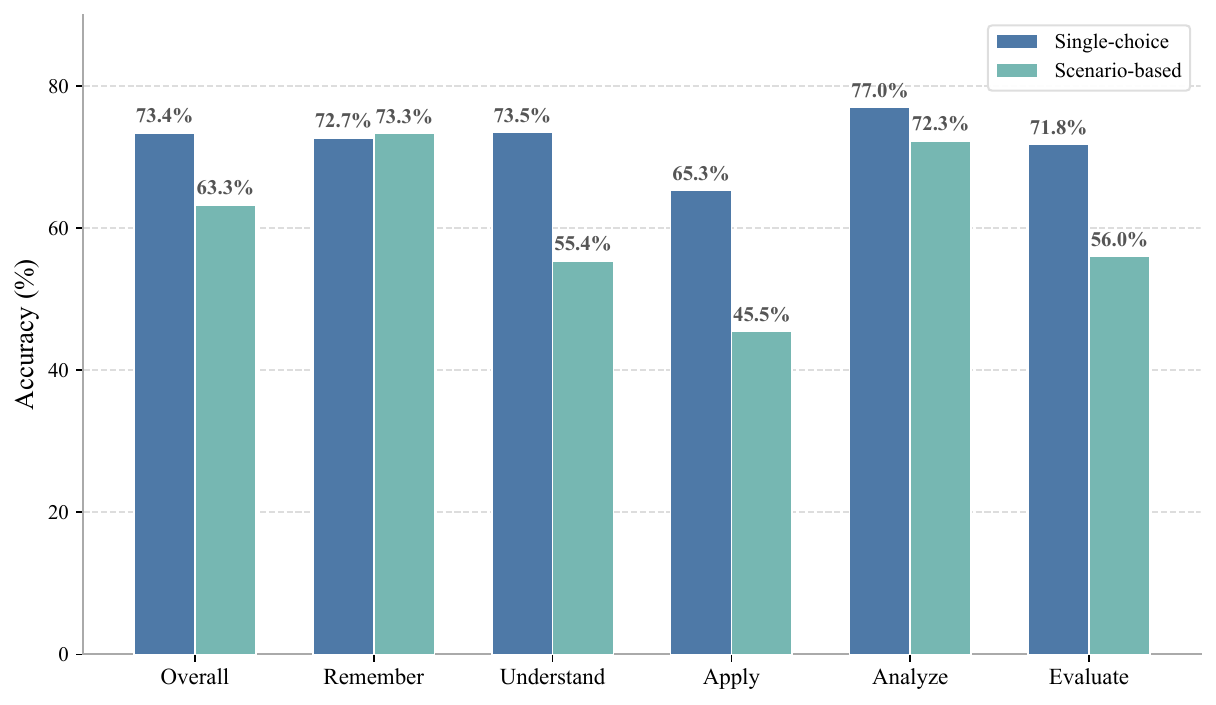}
  \caption{
  Baseline accuracy across question formats and cognitive levels. The model is evaluated on 811 single-choice questions and 379 scenario-based questions.
  }
  \label{fig:single_vs_scenario_accuracy}
\end{figure}

As shown in Figure~\ref{fig:single_vs_scenario_accuracy}, the baseline experiments are conducted on 811 single-choice questions and 379 scenario-based questions. The overall accuracy is 73.4\% for single-choice questions and 63.3\% for scenario-based questions. From the perspective of cognitive levels, the Apply level is the weakest category in both question formats, with accuracies of 65.3\% for single-choice questions and 45.5\% for scenario-based questions. This indicates that the model still has limited ability to transfer professional knowledge to concrete problem-solving contexts. In addition, the overall performance on scenario-based questions is substantially lower than that on single-choice questions, with more pronounced gaps in higher-order cognitive tasks.

\begin{figure}[!htbp]
  \centering
  \includegraphics[width=\linewidth]{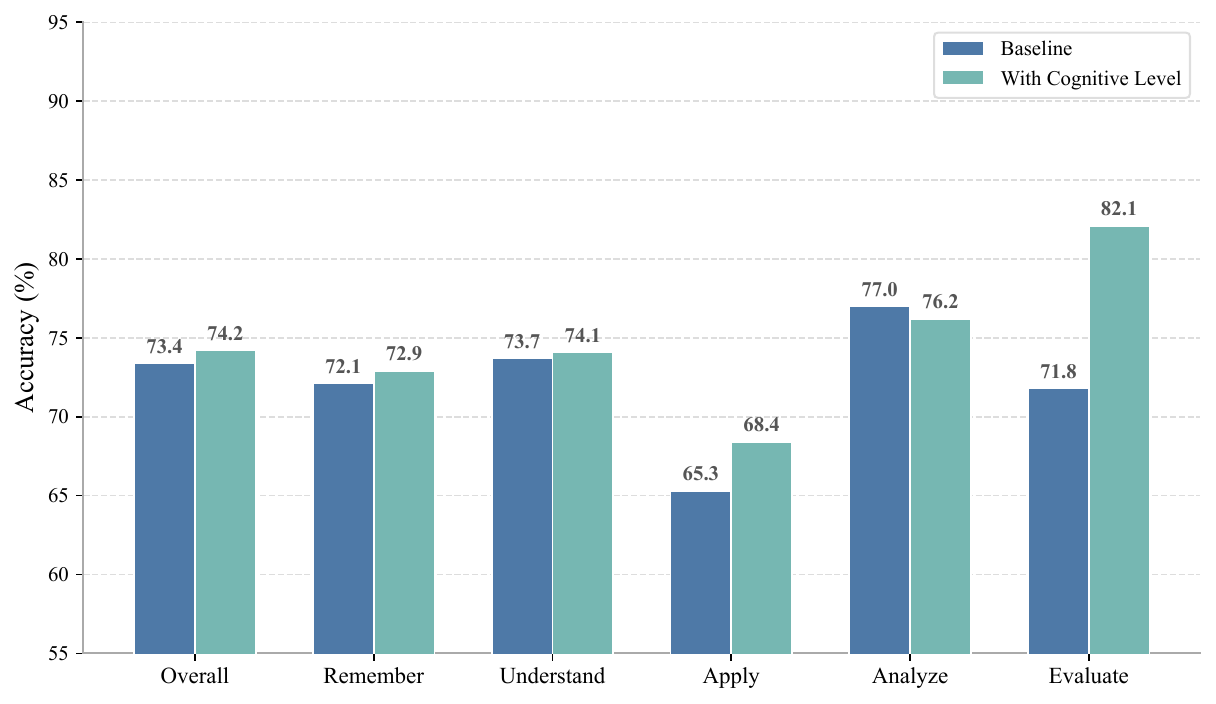}
  \caption{
  Effect of cognitive-level injection on single-choice questions. The baseline model is compared with a variant that explicitly includes the cognitive level in the prompt.
  }
  \label{fig:baseline_vs_cognitive_level}
\end{figure}

Figure~\ref{fig:baseline_vs_cognitive_level} further investigates whether cognitive-level information can serve as an effective control signal for model reasoning. After injecting the cognitive level into the prompt, the overall accuracy increases from 73.4\% to 74.2\%. Although the overall improvement is moderate, the gains are more evident for several cognitively demanding categories. In particular, the accuracy on Apply questions improves from 65.3\% to 68.4\%, and the accuracy on Evaluate questions increases from 71.8\% to 82.1\%. These results suggest that cognitive-level information can help the model adjust its reasoning strategy according to the required cognitive demand, especially for questions involving knowledge application and judgment.

% \begin{figure}[htbp]
%   \centering
%   \includegraphics[width=\linewidth]{Fig/prompt_ablation_accuracy.pdf}
%   \caption{
%   Comparison of different prompt-level cognitive guidance strategies. Level Only denotes injecting only the cognitive-level label, while Level + Description additionally includes a textual description of the cognitive level.
%   }
%   \label{fig:prompt_ablation_accuracy}
% \end{figure}

\begin{figure}[!htbp]
  \centering
  \includegraphics[width=0.72\linewidth]{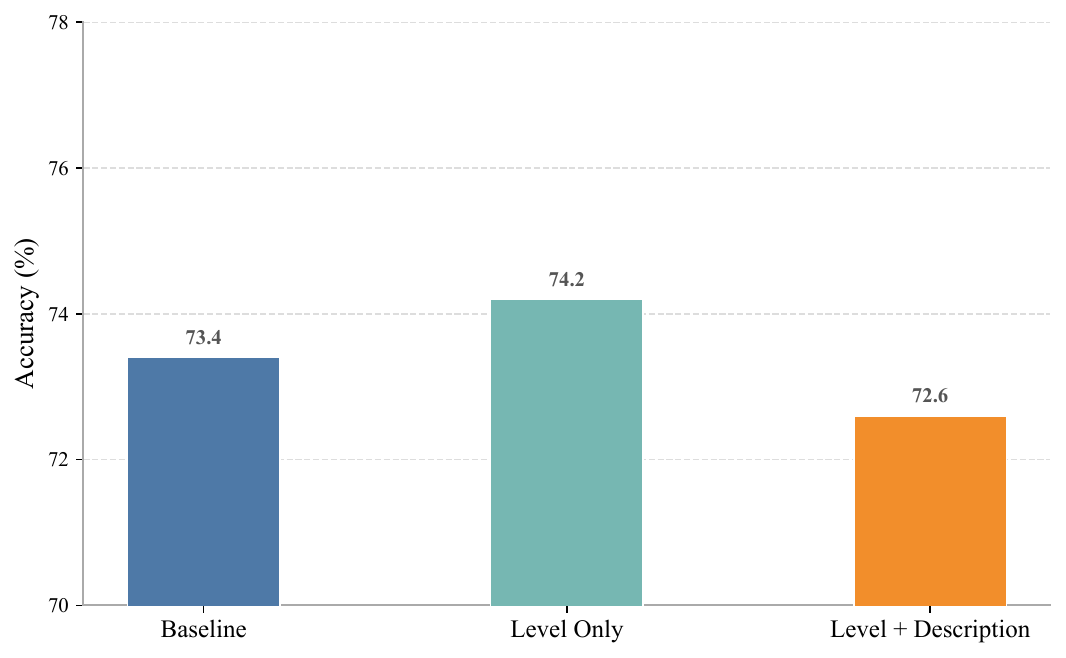}
  \caption{Comparison of prompt-level cognitive guidance strategies. Level Only injects only the cognitive-level label, whereas Level + Description additionally includes its textual description.}
  \label{fig:prompt_ablation_accuracy}
\end{figure}

To further analyze the form of cognitive guidance, Figure~\ref{fig:prompt_ablation_accuracy} compares three prompt settings: the baseline prompt, a prompt with only the cognitive-level label, and a prompt with both the cognitive-level label and its textual description. Injecting only the cognitive-level label improves accuracy from 73.4\% to 74.2\%, confirming that even lightweight cognitive guidance can benefit professional QA. However, adding a textual description of the cognitive level decreases accuracy to 72.6\%. This suggests that overly verbose cognitive descriptions may introduce additional prompt noise or distract the model from task-relevant evidence. Therefore, in the subsequent CogRAG framework, we use cognitive levels as compact control signals rather than relying on lengthy natural-language descriptions.

\section{Selection of the Retrieval Baseline}
\label{secA3}
%%%%%%%% 检索基准的选择

In terms of retrieval methods, we further compare BM25, Dense, and Hybrid retrieval under the Qwen3-8B single-choice setting. As shown in Table~\ref{tab:retrieval_baselines}, Dense retrieval achieves the best overall performance, reaching an accuracy of 78.2\%, which improves the baseline by 4.8 percentage points. It also brings consistent improvements across all cognitive levels, with the most pronounced gain observed at the Remember level, where accuracy increases from 72.7\% to 80.6\%. In contrast, BM25 achieves an overall accuracy of 72.1\%, slightly lower than the baseline, while Hybrid retrieval obtains 71.9\%, also failing to outperform Dense retrieval. These results indicate that, in the current nutrition exam setting, Dense retrieval is more effective than BM25 in covering paraphrased expressions and semantically relevant content, while simply combining sparse signals does not provide additional benefits. Therefore, we use Dense retrieval as the base retrieval method for CogRAG in the subsequent experiments.

\begin{table}[!htbp]
\centering
\small
\caption{Comparison of retrieval methods under the Qwen3-8B single-choice setting.}
\label{tab:retrieval_baselines}
\begin{tabular}{l c c c c c c c}
\toprule
\textbf{Method} & \textbf{Overall} & \textbf{Macro} & \textbf{Rem.} & \textbf{Und.} & \textbf{App.} & \textbf{Ana.} & \textbf{Eva.} \\
\midrule
Baseline & 73.4 & 72.0 & 72.7 & 73.5 & 65.3 & 77.0 & 71.8 \\
BM25     & 72.1 & 71.0 & 69.8 & 72.7 & 66.3 & 74.6 & 71.8 \\
Dense    & 78.2 & 77.8 & 80.6 & 77.1 & 74.5 & 79.8 & 76.9 \\
Hybrid   & 71.9 & 70.0 & 69.8 & 73.4 & 62.2 & 75.4 & 69.2 \\
\bottomrule
\end{tabular}
\end{table}

\end{appendices}

\section*{ORCID}
\noindent Xudong Wang: \url{https://orcid.org/0009-0005-3372-566X}\\
Zilong Wang: \url{https://orcid.org/0000-0002-7728-2056}\\
Su Kui: \url{https://orcid.org/0000-0002-7152-7841}\\
Zhaoyan Ming: \url{https://orcid.org/0000-0002-6766-4579}

\bibliographystyle{ws-jcsc}
\bibliography{main}

\end{document}